\PassOptionsToPackage{pdfpagelabels=false,hyperfootnotes=false}{hyperref}
\documentclass{IOS-Book-Article}

\usepackage{mathptmx}
\usepackage[T1]{fontenc}
\usepackage{amsmath}
\usepackage{booktabs}
\usepackage{tabularx}
\usepackage{graphicx}
\usepackage{placeins}
\graphicspath{{figures/}}
\usepackage{url}
\hypersetup{pageanchor=false}

\begin{document}

\pagestyle{headings}
\def\thepage{}
\begin{frontmatter}

\title{Where Do Legal Factors Live in LLMs?}

\markboth{}{}

\author[A]{\fnms{Li} \snm{Zhang}%
\thanks{Corresponding Author: Li Zhang; e-mail: liz239@pitt.edu.}}
and
\author[A]{\fnms{Kevin~D.} \snm{Ashley}}

\runningauthor{L. Zhang and K.D. Ashley}
\address[A]{University of Pittsburgh, USA}

\begin{abstract}
In case-based legal reasoning, legal professionals identify legally
significant fact patterns in a new dispute and compare them with those of
decided cases. In AI \&
Law, HYPO dimensions and CATO factors represent these patterns symbolically,
organizing expert knowledge distilled from case narratives as explicit,
structured, interpretable, and auditable legal concepts and relations that
support case comparison, legal argument, and precedent-based reasoning.
Large language models (LLMs) are increasingly used for legal tasks,
yet their internal organization of legal knowledge remains difficult to trace.
We investigate whether LLMs form internal representations corresponding to all
26 CATO factors. Two research questions examine linear decodability and
sparse localization with sparse autoencoders. Experiments cover three open
models, Qwen3-1.7B, Qwen3-8B, and Gemma-2 2B, supported by cross-model
comparison and human audit of controlled natural-language passages. Results
show that all 26 factors are linearly decodable in all three models.
Single-feature sparse readouts cover the complete factor catalog, with
best-layer test AUCs ranging from 0.86 to 1.00. Five selected features
approach full-vector readout performance. We report 78 model-specific
layer--feature coordinates, one for each factor in each model. Together,
these results show that information corresponding to expert-defined legal
factors can be read from the internal activations of LLMs processing legal
narratives. By locating these
readouts, our study connects legal language and the symbolic abstractions of
HYPO and CATO to learned internal representations.
\end{abstract}

\end{frontmatter}
\begingroup
\renewcommand{\thefootnote}{}
\footnotetext{Code and corpus: \url{https://github.com/lizhang-AIandLaw/FactorScope-LLM}.}
\endgroup
\markboth{}{}

\section{Introduction}\label{sec:intro}

Case-based legal reasoning compares a new case with decided cases. Legal
professionals identify the facts that made a difference in earlier cases,
argue from similarities and differences, and explain outcomes in those terms.
AI \& Law research has long represented such knowledge symbolically, at a
level between raw case facts and ultimate legal
conclusions~\cite{ashley1990,aleven1997}. HYPO and
CATO are classic examples. HYPO represents factual differences that favor
different parties in trade-secret disputes as dimensions~\cite{ashley1990};
CATO organizes such patterns into 26 \emph{factors}, stereotypical fact
patterns distilled by legal experts from case narratives~\cite{aleven1997}.
Factors make legally significant differences explicit, supporting case
comparison, argument construction, and explanations of which facts support a
conclusion~\cite{bruninghaus2003,ashley2009,horty2012,benchcapon2003,gray2023}.

LLMs can identify factors and perform other legal tasks~\cite{gray2023,
savelka2023,guha2023,katz2024,blairstanek2023,chalkidis2020}. Their outputs,
however, leave open how legal-factor information is represented internally.
For example, a model may associate a passage with reverse engineering because
it recognizes that a product's information can be recovered by inspection,
because it recognizes familiar wording, or because it recognizes the surrounding
commercial context. Examining internal representations could help diagnose
such differences and provide evidence for auditing legal applications.

\begin{figure}[t]
\centering
\includegraphics[width=\linewidth]{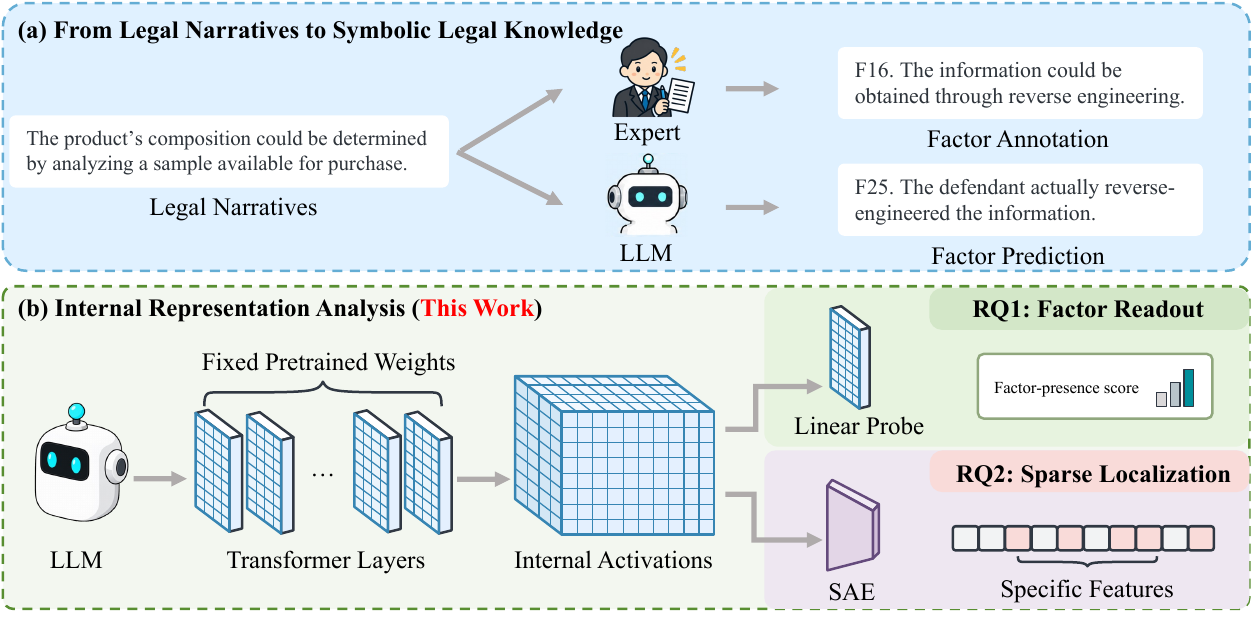}
\caption{Overview of legal-factor analysis. (a) HYPO and CATO employ legally
significant fact patterns abstracted from natural-language narratives.
Subsequent work prompts LLMs to identify these factors in text.
(b) This work examines internal representations in pretrained LLMs:
RQ1 reads factor presence from full activations, and RQ2 identifies one or
five SAE features supporting the readout.}
\label{fig:conceptual}
\end{figure}

\begin{samepage}
We ask: \emph{Can internal signals corresponding to expert-defined legal
factors be identified in LLMs?} CATO supplies the legal reference against which
these signals can be examined. We address two research questions:
\begin{description}
\item[RQ1.] Can the presence of each CATO factor be read from the model's
internal activations by a simple classifier?
\item[RQ2.] Can this information be localized to a small number of
inspectable internal features?
\end{description}
\end{samepage}
For RQ1, we train a linear \emph{probe}: a simple classifier that reads the
internal state of an LLM. For RQ2, we use a sparse autoencoder (SAE) to
express that state in terms of candidate features, then test how well one or
five selected features distinguish factor-present from factor-absent passages.
We report concrete layer and feature coordinates for every factor in each
model. The LLMs and released SAEs remain fixed; only the separate
readout classifiers are fitted to factor labels. The reported coordinates
give legal experts and developers concrete objects to inspect. A readout can
be tested on new passages that experts have labeled against the factor
definitions, and the activations of the selected features on individual
passages can be compared with those definitions.

Our experiments cover the complete 26-factor catalog and three model--SAE
pairs: Qwen3-1.7B, Qwen3-8B, and Gemma-2 2B. We construct controlled passages,
validate their labels across models and through human audit, and evaluate
factor readouts on a separate test partition. All 26 factors are readable from the
internal states of all three models, and a few SAE features provide strong
readouts. Factor information is thus readable both from full activation
vectors and from a small set of identifiable SAE features.
Figure~\ref{fig:conceptual} situates our study in this line of work.

\section{Background and Related Work}\label{sec:background}

\subsection{Legal Factors as a Reference for Internal Representations}
A CATO factor is a defeasible, stereotypical fact pattern that favors one
side~\cite{aleven1997}, descended from HYPO's dimensions~\cite{ashley1990,
rissland2002}. It abstracts a legally significant pattern from a narrative and
links it to one side's claim. This intermediate level, between raw facts and
ultimate issues, supports models of precedential constraint~\cite{horty2012,
horty2011}, theory construction with values~\cite{benchcapon2003,chorley2005},
and dialogue over precedents~\cite{prakken1998}. Table~\ref{tab:factors} lists
the catalog used throughout this study.

\begin{table}[t]
\caption{The CATO factor catalog (F1--F27; F9 is vacant). (P)/(D) marks the
party the factor favors.}
\label{tab:factors}
\centering
\footnotesize
\setlength{\tabcolsep}{3pt}
\begin{tabularx}{\textwidth}{@{}l >{\raggedright\arraybackslash}p{40mm} X@{}}
\toprule
\textbf{ID} & \textbf{Factor (side)} & \textbf{Meaning (short)}\\
\midrule
F1 & Disclosure-in-negotiations (D) & Plaintiff disclosed its information to defendant in negotiations.\\
F2 & Bribe-employee (P) & Defendant paid plaintiff's employee to defect with its information.\\
F3 & Employee-sole-developer (D) & Defendant's employee solely developed plaintiff's product.\\
F4 & Agreed-not-to-disclose (P) & Defendant signed a nondisclosure agreement with plaintiff.\\
F5 & Agreement-not-specific (D) & The agreement did not specify what was confidential.\\
F6 & Security-measures (P) & Plaintiff adopted concrete security measures.\\
F7 & Brought-tools (P) & Plaintiff's ex-employee brought development materials along.\\
F8 & Competitive-advantage (P) & Access to plaintiff's information saved defendant time or expense.\\
F10 & Secrets-disclosed-outsiders (D) & Plaintiff disclosed product information to outsiders.\\
F11 & Vertical-knowledge (D) & The information concerns customers and suppliers.\\
F12 & Outsider-disclosures-restricted (P) & Disclosures to outsiders carried confidentiality restrictions.\\
F13 & Non-competition-agreement (P) & The parties entered a non-competition agreement.\\
F14 & Restricted-materials-used (P) & Defendant used materials under confidentiality restrictions.\\
F15 & Unique-product (P) & Plaintiff was the only manufacturer of the product.\\
F16 & Info-reverse-engineerable (D) & The information could be learned by reverse-engineering.\\
F17 & Info-independently-generated (D) & Defendant developed its product independently.\\
F18 & Identical-products (P) & Defendant's product was identical to plaintiff's.\\
F19 & No-security-measures (D) & Plaintiff adopted no security measures.\\
F20 & Info-known-to-competitors (D) & The information was known to competitors.\\
F21 & Knew-info-confidential (P) & Defendant knew the information was confidential.\\
F22 & Invasive-techniques (P) & Defendant used invasive techniques to obtain the information.\\
F23 & Waiver-of-confidentiality (D) & Plaintiff waived confidentiality by agreement.\\
F24 & Info-obtainable-elsewhere (D) & The information was available from public sources.\\
F25 & Info-reverse-engineered (D) & Defendant actually reverse-engineered the information.\\
F26 & Deception (P) & Defendant obtained the information through deception.\\
F27 & Disclosure-in-public-forum (D) & Plaintiff disclosed the information in a public forum.\\
\bottomrule
\end{tabularx}
\end{table}

Systems have used factors for outcome prediction~\cite{bruninghaus2003,
grabmair2017}, extraction from case text~\cite{ashley2009}, and legal
annotation~\cite{gray2023,savelka2023,zhang2026retrieval}. Broader legal
benchmarks assess task performance~\cite{guha2023,katz2024,chalkidis2020,
zhang2025llms,zhang2026dlawbench,shi2026plawbench}. Our study examines the
internal signals associated with these factors. For example, F16
\emph{info-reverse-engineerable} concerns whether information \emph{could} be
obtained by reverse engineering; F25 concerns whether the defendant actually
reverse-engineered it. The factor definitions specify the legal criterion a
candidate signal should capture.

\subsection{From Internal Activations to SAE Features}
\textbf{Activations and probes.} As an LLM reads a passage, each layer
produces a vector of numerical activations for each token. We examine the
\emph{residual stream}, the vectors that successive transformer blocks
update~\cite{elhage2021}. A linear probe learns a weighted sum of these
numbers to distinguish, for example, F16-present from F16-absent passages.
Only the separate classifier is trained. Successful
prediction on passages excluded from probe fitting shows that factor information is accessible in
the internal state~\cite{alain2016,conneau2018}.

\textbf{Sparse autoencoders.} Individual activation coordinates can mix
several kinds of information. An SAE reconstructs an activation vector using
a sparse combination of learned features~\cite{cunningham2023,bricken2023,elhage2022}.
Each feature has an activation strength for a given input; relatively few are
active for each token. We use publicly released SAEs
and test which features predict the presence of each legal factor.

We use passage-level factor labels to select features and fit separate
classifiers. We call a classifier based on selected feature activations a
\emph{readout}. It learns whether higher or lower activation predicts factor
presence, combining evidence when multiple features are used. A feature is a
learned direction in activation space.
Each model and layer has its own SAE; a feature ID identifies a direction
within that particular SAE. Labels assigned to our generated passages using
CATO's factor definitions provide the reference for evaluating these readouts.

For example, an SAE may contain 65{,}536 features; a one-feature readout
uses the activation strength of one selected feature, and a five-feature
readout combines five. We test these classifiers on passages not used to fit
them.

Public SAEs support this analysis for Qwen and Gemma~\cite{gao2024,
qwenscope2026,rajamanoharan2024,lieberum2024}, following work on sparse
feature readouts~\cite{gurnee2023}. The features used to represent information
can differ between SAE decompositions~\cite{chanin2024}. We therefore check
whether resampling changes the selected features and compare models.
Control tasks and evaluation sets help assess what probes
learn~\cite{hewitt2019,ravichander2021}. We test predictions on new passages
and compare selected features with randomly chosen ones.

\textbf{Reading the scores.} We report the area under the receiver operating
characteristic curve, abbreviated as AUC throughout. Because each factor task
has many more negatives than positives, we use AUC, which does not depend on
a decision threshold or on the class ratio. It measures how well a
readout ranks factor-present passages above factor-absent passages: 0.5
indicates chance-level ranking and 1 indicates perfect separation. For F16,
for example, a high AUC means that the readout consistently assigns higher
factor-presence scores to passages expressing reverse-engineerability than
to the negative examples. AUC summarizes ranking across decision thresholds.

\section{A Controlled Corpus for Factor Probing}\label{sec:data}

Real cases often contain several factors together. To examine one factor at a
time, we generate short passages for each factor and check that they express
it. We vary their business settings and wording. The corpus covers all 26
CATO factors and lets us compare classifiers on the same passages.

All passages are labeled using the same factor definitions, regardless of
their setting or wording.

\textbf{Generation and labels.} Tencent Hy3~\cite{hy3} was the primary
generator of two- to four-sentence passages.
DeepSeek-V4-Flash~\cite{deepseekv4flash}
independently annotated each passage against all 26 factor definitions;
Anthropic Haiku~4.5~\cite{anthropic2025haiku} adjudicated disagreements.
We retained a passage only when its target factor was confirmed and any
additional factor followed from the target under the catalog's entailment
relations. The resulting corpus contains 6{,}992 factor-present passages
(217--324 per factor) and 901 factor-free passages.
Table~\ref{tab:corpus_examples} illustrates the variation built into the corpus:
industry, information type, and narrative style change while the target
factor remains F19.

\begin{table}[ht]
\setlength{\tablewidth}{\textwidth}
\caption{Corpus examples illustrating variation in industry, information type, and narrative style. All four passages express F19; labels highlight changes relative to Example A.}
\label{tab:corpus_examples}
\centering
\footnotesize
\setlength{\tabcolsep}{3pt}
\begin{tabularx}{\textwidth}{@{}>{\raggedright\arraybackslash}p{28mm} X@{}}
\toprule
\textbf{Example} & \textbf{Corpus excerpt}\\
\midrule
\textbf{Example A}\newline Supplier network & BlueFern Genetics developed a supplier network for its biotech reagent line. The company kept the specifications on a shared drive with no access controls.\\
\addlinespace[3pt]
\textbf{Example B}\newline Information type & Tessara Analytics developed a formula for industrial chemicals used in polymer curing. The company kept the specifications on a shared drive with no access controls.\\
\addlinespace[3pt]
\textbf{Example C}\newline Industry & Almara Labs maintained a supplier network for its aerospace components on a shared drive with no access controls. The company did not restrict who inside the organization could open the files.\\
\addlinespace[3pt]
\textbf{Example D}\newline Narrative style & Complaint excerpt: Vantor Gaming is said to mirror Veltrix Systems' textiles pricing steps. The plaintiff took no steps to limit which employees could view the process.\\
\bottomrule
\end{tabularx}
\end{table}

\textbf{Human audit.} A random sample of 220 passages was checked against
the factor definitions with the model labels hidden: eight passages per
factor (208), interleaved with twelve factor-free controls. Human--LLM
agreement was 98.6\% (217/220), including all twelve controls.

\textbf{Training and evaluation.} The training partition contains 3{,}485
factor-present and 447 factor-free passages (3{,}932 in total); the test
partition contains 3{,}507 and 454 (3{,}961). Both contain varied narratives from the same generation
and validation procedure. We keep jointly generated passages together when
splitting the corpus, with no identical passages shared between partitions.
The test set is used for readout comparison and layer selection.

A TF-IDF classifier uses individual words and adjacent word pairs to predict
factor presence. It provides a comparison with classifiers that use the
model's internal activations.

\section{Probing Design}\label{sec:method}

RQ1 and RQ2 use the same corpus, labels, and train/test partitions. We average
each passage's activations over tokens, excluding the first
token (Figure~\ref{fig:conceptual}(b)). ``Full activations'' comprises all dimensions
of this residual vector at one sampled layer.

\textbf{Models and SAE settings.} We analyze three publicly released
pretrained base checkpoints: Qwen3-1.7B-Base, Qwen3-8B-Base, and Gemma 2 2B
(pretrained).\footnote{\raggedright Hugging Face model IDs: \path{Qwen/Qwen3-1.7B-Base},
\path{Qwen/Qwen3-8B-Base}, and \path{google/gemma-2-2b}. SAE repositories:
\path{Qwen/SAE-Res-Qwen3-1.7B-Base-W32K-L0_100},
\path{Qwen/SAE-Res-Qwen3-8B-Base-W64K-L0_100}, and
\path{google/gemma-scope-2b-pt-res}. Gemma checkpoints:
\path{layer_6/width_16k/average_l0_70},
\path{layer_13/width_16k/average_l0_84}, and
\path{layer_20/width_16k/average_l0_71}, each containing \path{params.npz}.} We refer to them as
Qwen3-1.7B, Qwen3-8B, and Gemma-2 2B below, and read the residual stream at
approximately one-quarter, one-half, and three-quarters of each layer stack.
Qwen3-1.7B and Qwen3-8B use Qwen-Scope TopK-100 residual
SAEs~\cite{qwenscope2026}: 32{,}768-dimensional at layers 7/14/21 of 28 and
65{,}536-dimensional at layers 9/18/27 of 36, respectively.
Gemma-2 2B uses 16{,}384-dimensional JumpReLU Gemma Scope SAEs at layers
6/13/20 of 26~\cite{lieberum2024}.

\textbf{RQ1: Classification and layer selection.} For each factor, positives
are its validated passages. Within each partition, negatives combine
factor-free passages and positives for other factors, excluding factors
known to entail the target.
For RQ1, we fit an unweighted $L_2$-regularized logistic classifier to the
mean-pooled residual activations at each sampled layer. For each factor, we
report the maximum raw-residual AUC across the sampled layers on the test set.

For F16, negatives include neutral business passages and passages about other
factors. We exclude passages whose factors imply F16, because they also satisfy
F16 and should not be treated as negatives.

\textbf{RQ2: Selecting a few features.} Within a layer, features are ranked
on passages used for readout fitting by $|\mathrm{AUC}-0.5|$: features rank
higher when their activations separate positives from negatives in either
direction. Constant features receive 0.5. We freeze the top $k\in\{1,5\}$ features
and fit a new unweighted $L_2$ readout to each set. ``Top-1'' and ``top-5''
therefore mean classifiers restricted to one and five selected features.
Twenty draws of five uniformly sampled SAE features, including inactive
ones, supply a random-selection reference. We retain results from all sampled
layers. For the direct comparison in Figure~\ref{fig:replication}, all four
readout types use the middle sampled layer (14, 18, and 13, respectively),
and each factor contributes equally to the mean AUC. We also summarize
per-factor maxima across layers for each readout size.
For the location inventory, we report each factor's layer with the highest top-1 test AUC
and the feature ID selected on the readout-fitting data. This same rule applies to all three
models.

At each reported location, we average the selected SAE feature's activation
over the passage's tokens. The classifier uses this average to score whether
the factor is present.

For Qwen3-8B, we resample the training passages with replacement three times,
keeping the sample size fixed, and reselect five features each time.
We compare the selected sets using mean pairwise Jaccard overlap: the number
of shared features divided by the number in either set. The pre-specified
stability threshold is 0.6. For four factors below it, ten bootstrap samples
test whether the selected features and their classifier weights remain similar.

\section{Results and Discussion}\label{sec:results}

\subsection{RQ1: Factor Information Is Readable from Internal Activations}
For all 26 factors in each of the three models, a linear classifier separates
factor-present from factor-absent passages with maximum held-out AUC
0.998--1.000. These predictions use the models' internal activations for the
test passages, showing that they contain information about each factor's
presence.

TF-IDF also achieves near-perfect AUC on the test set: words and word pairs
alone can distinguish these positive and negative passages, so RQ1 shows that
factor information is readable without separating it from the wording that
expresses it. RQ2 next asks whether a few SAE features can also distinguish
them.

\subsection{RQ2: A Few SAE Features Provide Strong Factor Readouts}
\begin{figure}[t]
\centering
\includegraphics[width=0.50\linewidth]{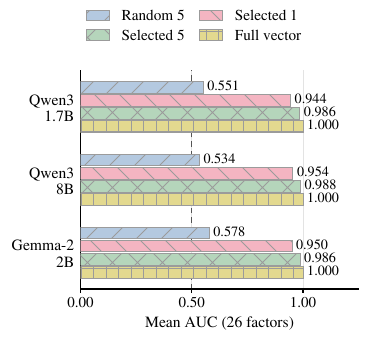}
\caption{Mean AUC across 26 factors at each model's middle sampled layer
(14/18/13). Random-5 averages 20 draws; features are selected on readout-fitting data.
The dashed line marks chance-level ranking.}
\label{fig:replication}
\end{figure}
Figure~\ref{fig:replication} directly compares random and selected SAE
features with the full activation vector. Within each model, all four readouts
use the same middle layer and the same 26 factor tasks. Random five-feature
sets achieve mean AUCs of 0.551, 0.534, and 0.578. A single selected feature
raises these means to 0.944, 0.954, and 0.950; five selected features reach
0.986, 0.988, and 0.986, close to the full-vector readouts. Thus, high
performance depends on which features are selected, and a few selected
features retain much of the factor-discriminating performance available in
the full activation vector.

In each model, every factor has a single-feature readout with AUC above 0.86
at one or more of the three tested layers. Taking each factor's best layer,
the median AUCs are 0.961, 0.972, and 0.961. For five-feature readouts, the
lowest of these per-factor best-layer AUCs are 0.965, 0.962, and 0.965.

\begin{figure}[t]
\centering
\includegraphics[width=\linewidth]{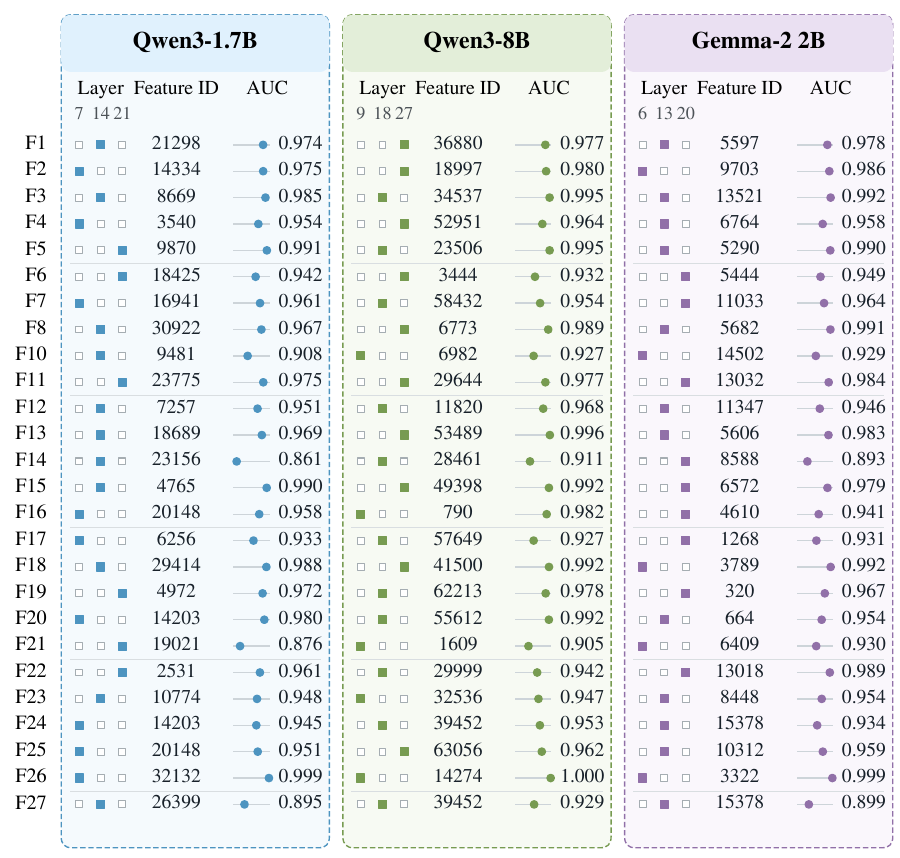}
\caption{Selected layer--feature locations for all 26 factors.
In each model panel, the filled square identifies the layer with the highest
test single-feature AUC; the adjacent number is the feature ID within that
layer's SAE. Dots mark AUC on horizontal tracks running from 0.85 (left)
to 1.00 (right), with exact values printed alongside.}
\label{fig:locations}
\end{figure}

Figure~\ref{fig:locations} gives concrete locations for all 26 factors in all
three models, together with AUC at each location. Each entry is the strongest
single-feature readout across the sampled layers. For F16, they are layer 7 / feature 20148 in
Qwen3-1.7B, layer 9 / feature 790 in Qwen3-8B, and layer 20 / feature 4610 in
Gemma-2 2B.

\FloatBarrier

\subsection{Discussion}\label{sec:discussion}

RQ1 establishes that factor information is readable from full activations;
RQ2 shows that a few selected SAE features retain much of this information.

\textbf{Layer distribution.} The selected locations span all three sampled
depths in every model. In Qwen3-1.7B, layers 7/14/21 supply the best top-1
readout for 9/11/6 factors; in Qwen3-8B, layers 9/18/27 supply 5/11/10; and
in Gemma-2 2B, layers 6/13/20 supply 5/12/9. The middle sampled layer is the
most frequent in each model, but fewer than half the factors select it.
Different factors are therefore best predicted at different layers.
The median AUC advantage over the next-best sampled layer is 0.011 in each
model, so another layer often gives a similarly strong prediction.

\textbf{Shared SAE features.} The 26 factors select 24 distinct layer--feature
coordinates in Qwen3-1.7B and 25 in each of Qwen3-8B and Gemma-2 2B.
In Qwen3-1.7B, F16 (reverse-engineerability) and F25 (actual reverse
engineering) both select layer 7 / feature 20148. F20 (information known to
competitors) and F24 (information obtainable elsewhere) share layer 7 /
feature 14203. In Qwen3-8B, F24 and F27 (public-forum disclosure) share layer
18 / feature 39452; in Gemma-2 2B, the same pair shares layer 13 / feature
15378.
This \emph{feature sharing} means that the activation of the same SAE feature
can help predict more than one legal factor.
For legal analysis, the relationship between the paired factors matters.
F16 concerns the possibility of obtaining information by reverse engineering;
F25 concerns an action actually taken. Their shared coordinate in Qwen3-1.7B
suggests that one feature responds to both questions about the same activity.
Similarly, F24 and F27 concern different routes by which information becomes
accessible. Reading the coordinates alongside the factor definitions makes
these relationships visible and identifies specific feature activations to
examine when comparing passages that express the related factors.

\textbf{From one feature to a small set.} Using several features can improve
prediction compared with using just one. For
F14, single-feature AUCs at the listed locations are 0.861, 0.911, and 0.893
in model order. Using five selected features at those same layers raises them
to 0.970, 0.969, and 0.965; full-vector AUCs are all at least 0.9998.
Thus, several features can predict a factor better than one alone, and
different factors can use some of the same features.
The small readout also provides a manageable object for inspection. For any
passage, its prediction can be decomposed into the contributions of the
selected feature activations and their fitted weights. Comparing those
contributions across passages can show which signals raise or lower the
factor score.

\textbf{Stability across samples.} For Qwen3-8B, 22 of 26 factors meet the
threshold for overlap between the sets of five selected features.
For F2, F8, F11, and F24, ten rounds of resampling select different sets of
five, but only eight or nine distinct features in total for each factor.
After retraining, the classifiers retain high AUC and similar relative
feature weights (weight-vector cosine similarity 0.95--0.99).
Thus, the exact selection varies, but predictions and feature weighting
remain stable.

The factor catalog lets us ask the same concrete questions of each model:
which factors can be predicted from one feature, which benefit from several,
and which use the same features. We can then compare the layers and features
that predict each factor across models.
The common reference is the legal factor: feature numbers belong to each
model's own SAE. The recurrence of strong sparse readouts across the three
models shows that, in each learned feature space, a few features suffice to
predict the catalog's factors. The combination of common
factor definitions, passage labels, and reported coordinates supports two
levels of comparison: whether the same factor information is readable in
each model, and how compactly each model represents the signals used to
predict it.

\section{Conclusion}\label{sec:conclusion}

Our results show that information corresponding to the fact patterns
captured by CATO factors can be read from the internal activations of LLMs
pretrained on large text corpora. Across three models, RQ1 establishes that all 26 factors can be
read from full activation vectors. RQ2 shows that much of this discriminating
information is accessible through one or a few selected SAE features. The
78 reported layer--feature coordinates give these correspondences concrete
locations in each model. Together, the results identify compact internal
readouts for patterns that legal experts express as symbolic factors.
These readouts partly overlap, with related factors sometimes sharing an SAE
feature.

The study connects legal natural language, symbolic legal knowledge, and
LLM internal representations (Figure~\ref{fig:conceptual}) through a concrete
test: can a classifier predict an expert-defined factor from the activations
produced when a model reads a passage?
This makes it possible to study how concepts defined by legal experts are
represented in models trained on language. Legal experts can use the factor
definitions to judge what a passage expresses, then test
whether the reported features predict those judgments.

This approach gives symbolic legal knowledge a further role: it supplies
expert-defined distinctions for studying what models learn from language.
Although the models have different internal coordinates, the same factor
definitions let us compare which distinctions their representations preserve
and which signals overlap. The contribution
is a way to investigate learned representations through questions grounded
in legal analysis. This study is a proof of concept on controlled narratives.

\section{Limitations and Future Work}\label{sec:limitations}

\textbf{Scope of the texts and legal domain.} The evidence comes from
controlled English narratives about U.S. trade-secret factors. The corpus
supports comparable measurements for all 26 factors. Naturally occurring
cases contain longer narratives, co-occurring factors, disputed facts, and
information spread across evidence and procedural history. The next step is
to evaluate the frozen readouts on the full texts of judicial opinions and
case files, reporting performance separately for individual and co-occurring
factors. Testing other legal domains and languages would show whether their
legal concepts can also be predicted from a few SAE features. The three open
models, with 1.7B to 8B parameters, are small by current standards; larger
models remain to be tested.

\textbf{Generalization and annotation.} Training and test passages share a
controlled generation procedure. Some positive and negative passages differ in length,
and TF-IDF performs strongly. Further evaluation should match length
profiles, test more varied wording, and include contrastive passages that
change the legal facts rather than the wording, for example security measures
taken by someone other than the plaintiff, or access restrictions with no
legal significance. Reporting precision, recall, and F1 at fixed thresholds
would complement AUC. Labels received cross-model checks and a blind
human audit comparing human judgments with model labels. A further study
with multiple experts could measure agreement between annotators and examine
disagreements about the underlying fact patterns.

\textbf{From representation to use.} Future work can test whether factor
readouts help explain errors that experts find in generated case summaries
and comparisons. If a summary omits confidentiality measures, does the
readout also fail to detect that factor in the input? If a model confuses
whether information could be reverse-engineered with whether it actually
was, do the two factors rely on the same SAE features? Intervention
experiments that suppress or amplify a selected feature would test whether a
model's outputs depend on it.
These tests could help legal experts and developers identify facts that
models repeatedly miss or confuse and target them for evaluation and
improvement. They would give legal expertise a concrete role in evaluating
and overseeing tools for case comparison, explanation, and legal reasoning
in practice.

\newpage
\bibliographystyle{vancouver}
\bibliography{references}

\end{document}